\documentclass[letterpaper, 10 pt, conference]{ieeeconf}
\IEEEoverridecommandlockouts
\usepackage{cite}
\usepackage{amsmath,amssymb,amsfonts}
\usepackage[ruled,linesnumbered,noend,vlined]{algorithm2e}
\usepackage{graphicx}
\usepackage{float}
\usepackage{hhline}
\usepackage{hyperref}
\usepackage{multirow}
\usepackage{stfloats}
\usepackage{textcomp}
\usepackage{xcolor}

\SetAlgoSkip{}
\def\BibTeX{{\rm B\kern-.05em{\sc i\kern-.025em b}\kern-.08em
    T\kern-.1667em\lower.7ex\hbox{E}\kern-.125emX}}

\makeatletter
\def\footnoterule{%
  \kern-5pt
  \hbox to \columnwidth{\hfill\vrule width 1\columnwidth height 0.4pt\hfill}
  \kern4.6pt
}
\makeatother

\begin{document}


\title{\LARGE \bf
DART-VLN: Test-Time Memory Decay and Anti-Loop Regularization for Discrete Vision-Language Navigation
}

\author{
\authorblockN{Shaoheng Zhang$^{1}$, Zhichen Li$^{1}$, Guangfu Ma$^{1,*}$, and Jie Mei$^{1,*}$}
\authorblockA{
$^{1}$Harbin Institute of Technology, Shenzhen \\
Guangdong, China
}
\thanks{This work was supported by the Shenzhen Fundamental Research Program
under Grant GXWD20231129140908002.}%
\thanks{*Corresponding authors: Guangfu Ma ({\tt\small magf@hit.edu.cn}) and
Jie Mei ({\tt\small jmei@hit.edu.cn}).}%
\thanks{Code is available at
\url{https://github.com/Japluto/DART-VLN}.}%
}

\maketitle
\thispagestyle{empty}
\pagestyle{empty}

\begin{abstract}
Memory-based agents for discrete vision-language navigation (VLN) operate under partial observability and may exhibit systematic inference-time failures even with strong pretrained backbones. This paper addresses two recurring problems: stale historical evidence during memory readout and inefficient local backtracking during action selection. We present DART-VLN, a training-free inference-time framework that combines Test-Time Memory Decay, which reweights stale and redundant memory slots without modifying their stored content, with Anti-Loop Regularization, a lightweight next-hop penalty that discourages immediate reversals. DART-VLN introduces no learnable parameters and leaves the navigation backbone unchanged. Experiments on R2R and REVERIE showed that memory decay preserved or improved task performance while reducing runtime. The addition of anti-loop regularization further shortened trajectories, reduced local backtracking, and achieved the best overall balance between navigation quality and efficiency among the evaluated GridMM variants. These results indicate that lightweight inference-time control can improve the reliability and efficiency of memory-based discrete VLN without retraining.
\end{abstract}



\section{Introduction}

Language-guided embodied navigation requires reliable sequential decision-making under partial observability. Vision-language navigation (VLN) requires an embodied agent to navigate an environment by grounding language instructions in visual observations \cite{anderson2018vision,zhang2024survey}. Among existing formulations, discrete VLN is especially attractive because it operates over explicit viewpoint graphs and supports controllable stepwise decision-making \cite{wang2023gridmm,chen2022duet,an2022bevbert}.

Recent advances in VLN have been driven by pretrained navigation models, explicit memory and map representations, larger training resources, and improved recovery and planning mechanisms \cite{hong2021vlnbert,guhur2021airbert,chen2021hamt,wang2023gridmm,gao2023zoneaware,dong2025sevln}. Together, these advances have substantially improved long-horizon reasoning and benchmark performance. Yet a practical gap remains at inference time: even strong memory-based discrete VLN backbones can behave unreliably when deployed with frozen parameters. Further gains often require retraining, architectural redesign, or heavier planning modules, making them less suitable when the goal is to improve a pretrained navigator without modifying its architecture or training procedure.

In this setting, two recurring failure modes become particularly important. The first appears at memory readout. Explicit navigation memory helps agents reason over longer trajectories by storing previously visited viewpoints, visual features, or map-level context \cite{wang2023gridmm,chen2021hamt,chen2022duet,an2022bevbert,gao2025gaussianmap,zhang2025cosmo,krantz2023iterative}. However, as navigation proceeds, stale or repeatedly observed evidence can persist in memory after its utility has diminished, introducing noise into memory aggregation at decision time. The second appears at action selection. Even with a strong frozen backbone, agents still exhibit inefficient local behaviors such as immediate reversals and short loops \cite{ma2019regretful,wang2023dreamwalker,xu2025navq}. These behaviors do not always prevent success, but they lengthen trajectories, waste steps, and increase runtime.

\begin{figure*}[t]
    \centering
    \includegraphics[width=1.0\textwidth]{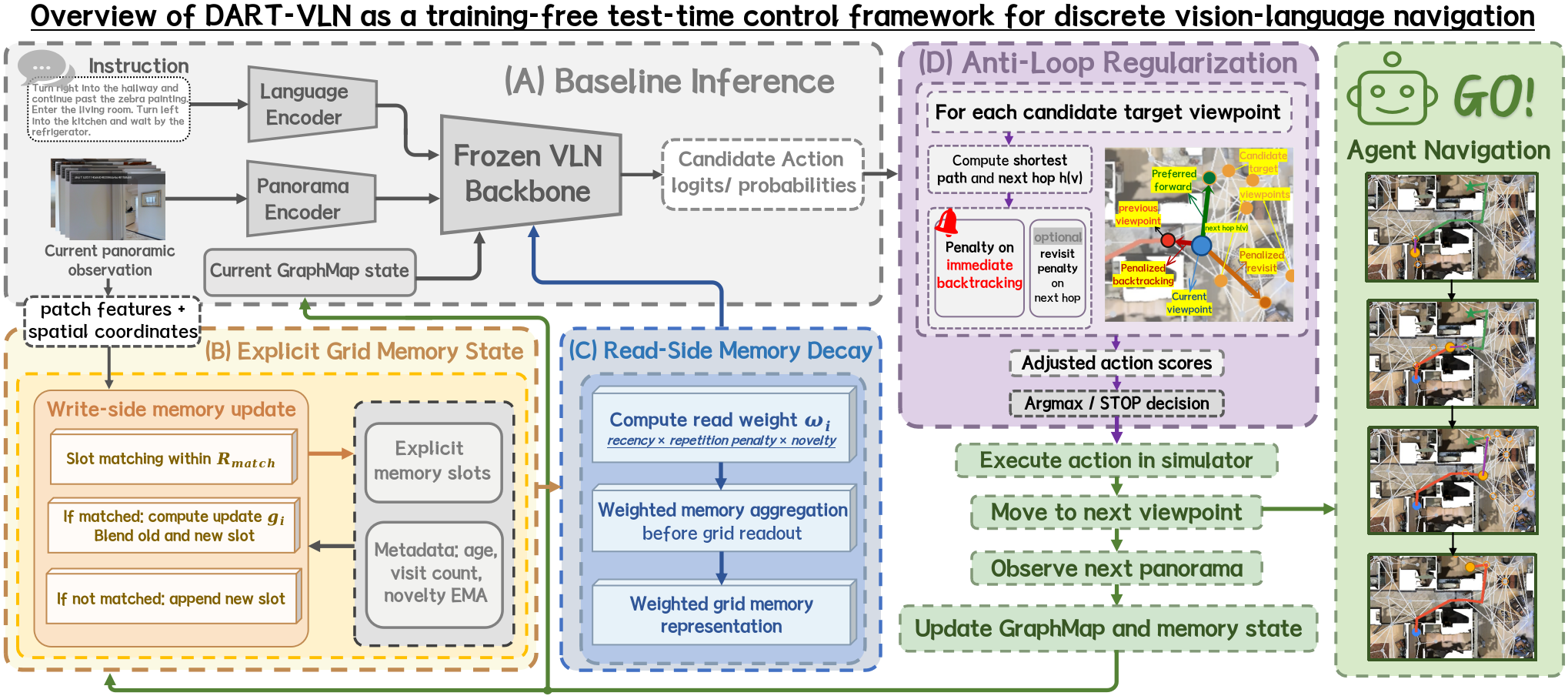}
    \vspace{-10pt}
    \caption{Overview of DART-VLN as a plug-in test-time control layer for discrete VLN pipelines with explicit memory. Memory decay reweights historical slots at readout, while anti-loop regularization applies next-hop penalties before argmax. The backbone remains frozen with no new learnable parameters.}
    \vspace{-15pt}
    \label{fig:framework}
\end{figure*}

This motivates a simple question: how much can be gained from lightweight test-time control alone, without retraining the model or redesigning the navigation stack? We answer this question with DART-VLN, a training-free test-time control framework for memory-based discrete VLN. As summarized in Fig.~\ref{fig:framework}, DART-VLN introduces two complementary mechanisms that operate directly inside the existing inference loop. \emph{Test-Time Memory Decay} is a read-side reweighting rule that suppresses stale and redundant evidence during memory aggregation without rewriting stored content. \emph{Anti-Loop Regularization} is a lightweight next-hop penalty that discourages immediate backtracking during action selection. Together, these components improve inference behavior while leaving the learned backbone untouched and introducing no new learnable parameters.

DART-VLN is designed as a plug-in test-time control layer for discrete VLN pipelines with explicit memory. In our experiments, we instantiate it on a GridMM-based navigator and evaluate it on R2R and REVERIE \cite{wang2023gridmm}. Across both benchmarks, read-side decay preserves or improves task performance, while anti-loop regularization reduces local reversals, trajectory length, and runtime. These results suggest that lightweight test-time control can strengthen memory-based discrete VLN without retraining.

\subsection{Contributions}
\begin{itemize}
\item We present DART-VLN as a training-free plug-in layer for discrete VLN pipelines with explicit memory, using read-side decay to suppress stale and redundant evidence without rewriting stored content.
\item We introduce a lightweight next-hop regularizer that discourages immediate backtracking and improves trajectory efficiency at inference time.
\item Experiments on R2R and REVERIE with a GridMM-based navigator show that read-side decay preserves or improves task performance, while combining it with anti-loop regularization offers the best navigation-efficiency balance among the evaluated variants.
\end{itemize}

\section{Related Work}

\textit{Memory and map-based discrete VLN:}
Explicit memory and map representations support long-horizon reasoning in discrete VLN by storing visited viewpoints, aggregating graph context, or building richer topological states \cite{wang2023gridmm,chen2022duet,an2022bevbert,georgakis2022crossmodal,gao2025gaussianmap,zhang2025cosmo}. As a plug-in layer, DART-VLN complements this line by controlling how existing memory is read at inference time, without introducing a new memory structure or map encoder.

\textit{Backtracking, rewind, and local recovery:}
A second line of work studies local navigation failures such as wrong turns, backtracking, and short loops \cite{ma2019regretful,wang2023dreamwalker,xu2025navq}. Regret-aware, foresighted, and planning-oriented methods show that recovery behavior matters for trajectory efficiency, but often rely on learned correction policies or heavier planning modules. Our method shares this motivation but implements it as a lightweight next-hop regularizer applied directly to local action scores.

\textit{Inference-time control versus retraining-heavy improvement:}
Strong VLN gains also come from architectural extensions, richer scene modeling, and heavier planning \cite{gao2023zoneaware,dong2025sevln}. Although effective, these approaches can be less suitable when the goal is to improve a deployed navigator without retraining. Recent studies have also explored training-free improvements to VLN at inference time \cite{rajabi2025travel,ko2025atena}; DART-VLN follows this direction through read-side memory control and action-side local regularization.

\section{Method}

\subsection{Problem Setting and Baseline Navigation Inference}

We consider a memory-based agent operating on a discrete navigation graph \(G=(V,E)\), where each node denotes a viewpoint and each edge represents a feasible transition. Given a natural-language instruction \(x\), the agent starts from an initial viewpoint \(v_0\) and makes sequential navigation decisions over the graph, producing a trajectory \(\tau=(v_0, v_1, \dots, v_T)\). At each step \(t\), the agent receives the current visual observation at \(v_t\), retrieves historical information from an explicit navigation memory, and scores the candidate actions reachable from the current viewpoint.

For evaluation, we instantiate DART-VLN on a GridMM-based discrete VLN pipeline \cite{wang2023gridmm}. The navigator encodes the instruction and current observation, aggregates historical evidence from explicit memory slots \(\{m_i\}\), and computes action scores \(s_t(v)\) for candidate target viewpoints \(v\). This explicit-memory design provides useful long-horizon context, but stale or repeatedly observed evidence may continue to influence the current decision.

More generally, DART-VLN requires only explicit memory readout and pre-selection candidate scores, enabling plug-in use without modifying the backbone. It reweights memory slots during aggregation and regularizes candidate scores before argmax decoding. The backbone remains frozen, and no learnable parameters are introduced.

Motivated by reliable embodied navigation under partial observability \cite{krantz2020vlnce}, DART-VLN is designed for memory-based discrete VLN pipelines but validated here only on a GridMM-based navigator. The following subsections describe its read-side memory control and action-side regularization.

\subsection{Test-Time Memory Decay}

\subsubsection{Motivation}

In memory-based navigation, stale or repeatedly observed slots may remain active after their usefulness has faded \cite{gao2025gaussianmap,zhang2025cosmo}. We therefore use a conservative read-side strategy: stored memory content is not rewritten, and only slot contributions are reweighted during aggregation.

\subsubsection{Slot Metadata}

For each memory slot \(m_i\), we maintain three lightweight metadata variables: slot age \(a_i\), visit count \(c_i\), and novelty \(n_i\). The slot age is defined as
\[
a_i = t - t_i^{\mathrm{last}},
\]
where \(t_i^{\mathrm{last}}\) is the most recent step at which slot \(i\) was refreshed. The visit count \(c_i\) records how often the corresponding region has been observed. The novelty term \(n_i\) captures recent feature change and is smoothed over time.

Concretely, let \(f_i^{\mathrm{old}}\) and \(f_i^{\mathrm{new}}\) denote the previous and current slot features. We first compute an instantaneous novelty score
\begin{equation}
\nu_i = \mathrm{clip}\big(1 - \cos(f_i^{\mathrm{old}}, f_i^{\mathrm{new}}),\, 0,\, 1\big),
\label{eq:novelty_score}
\end{equation}
and then update the novelty estimate with an exponential moving average:
\begin{equation}
n_i^{(t)} = \rho n_i^{(t-1)} + (1-\rho)\nu_i.
\label{eq:novelty_ema}
\end{equation}
Here, \(\rho \in [0,1)\) controls how strongly past novelty estimates are retained over time. In our implementation, \(\rho=0.5\). Together, these metadata provide a lightweight summary of whether a slot is recent, repeatedly observed, or still undergoing meaningful feature changes.

\subsubsection{Memory Reweighting}

Based on the metadata above, we assign a heuristic readout weight \(w_i\) to each memory slot. Since \(\nu_i\) is clipped to \([0,1]\) and \(n_i\) is obtained by exponential smoothing, we also have \(n_i \in [0,1]\). The readout weight is defined as
\begin{equation}
\begin{aligned}
w_i = \mathrm{clip}\Bigl(
&\exp(-\lambda a_i)
\Bigl(1-\alpha \frac{c_i}{c_i+1}\Bigr)
(0.5+0.5n_i),\\
& w_{\min},\, w_{\max}
\Bigr).
\end{aligned}
\label{eq:memory_decay}
\end{equation}

This formulation combines three factors. The recency term \(\exp(-\lambda a_i)\) downweights slots that have not been refreshed for many steps. The repetition term \(\bigl(1-\alpha \frac{c_i}{c_i+1}\bigr)\) reduces the influence of repeatedly observed regions, with \(\alpha\) controlling the suppression strength. The novelty term \((0.5+0.5n_i)\) assigns greater weight to slots showing recent feature changes. Finally, clipping to \([w_{\min}, w_{\max}]\) prevents overly weak or aggressive reweighting.

In the default DART-VLN configuration, these weights are applied only during memory readout. The weighted memory is then passed into the original navigation pipeline without modifying the stored slot content. This keeps the intervention training-free and straightforward to isolate in ablation studies.

\subsubsection{Boundary of Decay-Only}

In the \emph{decay-only} setting, only the read-side reweighting in Eq.~(\ref{eq:memory_decay}) is applied; stored slot representations and baseline memory updates remain unchanged. Thus, \emph{decay-only} is a read-side denoising rule rather than a learned memory update mechanism.

\subsubsection{Write-Side Memory Update Variants}

Beyond read-side decay, we test a more aggressive write-side update in which matched slots are refreshed by a heuristic gate based on novelty, age, and repetition, while unmatched observations are appended as new slots. We report this as \emph{update-only} and its combination with decay as \emph{full-mode}. Because these variants are less stable, we include them as stress tests of more aggressive memory intervention rather than as part of the default configuration.

\begin{algorithm}[t]
\caption{Default DART-VLN Inference}
\label{alg:dart_main}
\footnotesize
\DontPrintSemicolon
\SetKwInOut{Notation}{Notation}
\SetKwInOut{KwIn}{Input}
\SetKwInOut{KwOut}{Output}
\SetKwFor{ForEach}{for each}{do}{}
\SetKwFunction{DARTInfer}{DART-VLN-Infer}
\SetKwProg{Fn}{Function}{:}{}

\Notation{$h(v)$: graph next hop of candidate viewpoint $v$\;
$p_t(v)$: anti-loop penalty\;
$s'_t(v)$: adjusted action score\;}
\KwIn{Instruction $x$; initial viewpoint $v_0$; max step $T_{\max}$; memory slots $\{m_i\}$ with metadata $\{a_i,c_i,n_i\}$}
\KwOut{Navigation trajectory $\tau$}

\Fn{\DARTInfer{$x, v_0, \{m_i,a_i,c_i,n_i\}, T_{\max}$}}{
Initialize $\tau \leftarrow (v_0)$ and the visit counts from the start viewpoint\;
\For{$t\leftarrow 0$ \KwTo $T_{\max}-1$}{
Encode the instruction and current observation at $v_t$\;
Compute read-side memory weights $w_i$ using Eq.~(\ref{eq:memory_decay})\;
Aggregate weighted memory and obtain candidate scores $\{s_t(v)\}$\;
\ForEach{candidate target viewpoint $v$}{
Compute graph next hop $h(v)$ from the current viewpoint\;
Compute penalty $p_t(v)$ using Eq.~(\ref{eq:anti_loop_penalty})\;
Update adjusted score $s'_t(v)\leftarrow s_t(v)-p_t(v)$\;
}
Select the next action by $\arg\max_v s'_t(v)$\;
\If{STOP is selected}{
\KwRet $\tau$\;
}
Execute the transition and append the new viewpoint to $\tau$\;
Update visit counts, memory, and slot metadata for the next step\;
}
\KwRet $\tau$\;
}
\end{algorithm}

\begin{table*}[t]
\small
\setlength{\tabcolsep}{0.29cm}
\renewcommand{\arraystretch}{1.1}
\centering
\caption{Results on the R2R dataset}
\vspace{-8pt}
\label{tab:r2r_main}
\begin{tabular}{c|cccc|c|cccc|c}
\hline
\multirow{2}{*}{\textbf{Methods}} 
& \multicolumn{5}{c|}{\textbf{Val Unseen}} 
& \multicolumn{5}{c}{\textbf{Test Unseen}}\tabularnewline
\cline{2-11}
& TL$\downarrow$ & NE$\downarrow$ & SR$\uparrow$ & SPL$\uparrow$ & Runtime(s)$\downarrow$
& TL$\downarrow$ & NE$\downarrow$ & SR$\uparrow$ & SPL$\uparrow$ & Runtime(s)$\downarrow$\tabularnewline
\hline
VLNBERT~\cite{hong2021vlnbert} & 12.03 & 3.93 & 63 & 57 & -- & 12.37 & 4.11 & 63 & 57 & --\tabularnewline
AirBERT~\cite{guhur2021airbert} & 11.78 & 3.99 & 62 & 56 & -- & 12.40 & 4.15 & 62 & 57 & --\tabularnewline
SEvol~\cite{dong2025sevln} & 12.22 & 3.96 & 62 & 57 & -- & 13.42 & 4.17 & 62 & 57 & --\tabularnewline
HOP~\cite{qiao2023hopplus} & 12.28 & 3.81 & 64 & 57 & -- & 12.71 & 3.87 & 64 & 59 & --\tabularnewline
HAMT~\cite{chen2021hamt} & \textbf{11.44} & 2.31 & 66 & 61 & -- & 12.27 & 3.95 & 65 & 60 & --\tabularnewline
TD-STP~\cite{zhao2022tdstp} & -- & 3.23 & 70 & 63 & -- & -- & 3.72 & 67 & 61 & --\tabularnewline
DUET~\cite{chen2022duet} & 13.92 & 3.33 & 72 & 60 & -- & 14.71 & 3.63 & 69 & 59 & --\tabularnewline
BEVBert~\cite{an2022bevbert} & 14.52 & 2.83 & 74 & 64 & -- & 15.88 & 3.15 & 73 & 62 & --\tabularnewline
GridMM (baseline)~\cite{wang2023gridmm}  & 13.27 & 2.83 & 74 & 64 & 937.99 & 14.43 & 3.35 & 73 & 62 & 2312.52\tabularnewline
\hline
update-only & 13.92 & 2.86 & 75 & 63 & 891.12 & 16.2 & 3.37 & 72 & 59 & 1725.22\tabularnewline
decay-only & 13.29 & \textbf{2.59} & \textbf{76} & 65 & 742.61 & 14.52 & \textbf{3.19} & 73 & 62 & 1621.34\tabularnewline
full-mode & 13.77 & 2.80 & 75 & 64 & 1000.84 & 15.95 & 3.30 & 72 & 60 & 2475.63\tabularnewline
\textbf{decay+anti-loop} & \textbf{12.41$\downarrow$} & 2.69 & \textbf{76$\uparrow$} & \textbf{66$\uparrow$} & \textbf{666.37$\downarrow$} & \textbf{13.80$\downarrow$} & 3.38 & \textbf{74$\uparrow$} & \textbf{63$\uparrow$} & \textbf{1329.56$\downarrow$}\tabularnewline
\hline
\end{tabular}
\par\vspace{5pt}
\parbox{\textwidth}{\scriptsize \emph{Note:} Runtime is reported only for GridMM and its variants under the same implementation and hardware; cross-paper runtime is not directly comparable.}
\par\vspace{-12pt}
\end{table*}

\subsection{Anti-Loop Regularization for Navigation Decision-Making}

\subsubsection{Motivation}

Immediate backtracking and short loops waste steps and expose the agent to additional error-prone decisions \cite{ma2019regretful,wang2023dreamwalker,xu2025navq}. We therefore introduce a lightweight action-side regularizer that discourages local reversals while keeping the frozen policy otherwise unchanged.

\subsubsection{Next-Hop-Based Penalty}

Let \(s_t(v)\) denote the original action score assigned at step \(t\) to a candidate target viewpoint \(v\). Rather than penalizing the target directly, we examine its graph next hop \(h(v)\), the first local transition on the path from the current viewpoint to \(v\). We then define the penalty
\begin{equation}
p_t(v) =
\beta_{\mathrm{back}} \cdot \mathbb{I}[h(v)=v_{t-1}]
+
\beta_{\mathrm{rev}} \cdot \mathbb{I}[\mathrm{visit}(h(v)) \geq k],
\label{eq:anti_loop_penalty}
\end{equation}
and use it to adjust the action score:
\begin{equation}
s'_t(v) = s_t(v) - p_t(v).
\label{eq:anti_loop_adjusted}
\end{equation}

The first term penalizes a next hop that returns directly to the previous viewpoint \(v_{t-1}\). The second applies a smaller penalty to repeated revisits. Here, \(\mathrm{visit}(h(v))\) denotes prior visits to the next-hop viewpoint, \(\beta_{\mathrm{rev}}\) controls the revisit penalty, and \(k\) is its threshold. In the default setting, this term remains small and is applied only after repeated revisits, with \(k=2\). Immediate-backtracking suppression therefore remains the module's dominant effect.

This next-hop formulation is important because it targets the local transition actually executed at the current step, rather than only adjusting a more distant target score. It therefore aligns the regularizer with the trajectory-level behavior we aim to change.

\subsubsection{Inference Scope}

The anti-loop module is applied only at test time and targets deterministic, argmax-based navigation. In the default configuration, the anti-loop module operates alongside read-side memory decay and adjusts candidate scores before action selection. The module modifies neither the stop head nor the learned backbone and requires no additional planner. Its scope is deliberately narrow: it is a graph-local regularizer that improves trajectory efficiency with a frozen backbone. Because the penalty is finite and no action is masked, the agent can still backtrack when the original score provides sufficiently strong evidence.

\subsection{Design Principle: Conservative Test-Time Control}

DART-VLN follows a conservative test-time control principle: read-side reweighting suppresses stale evidence without perturbing stored representations, and the action-side regularizer discourages wasteful local loops without acting as a planner. We therefore treat the more aggressive write-side variants as stress tests rather than components of the default configuration. Algorithm~\ref{alg:dart_main} summarizes the resulting inference flow.

\section{Experiments}

\subsection{Experimental Setup}

We evaluate DART-VLN on Room-to-Room (R2R) \cite{anderson2018vision} and REVERIE \cite{qi2020reverie}. For R2R, Table~\ref{tab:r2r_main} reports TL, NE, SR, SPL, and runtime on \emph{val unseen} and \emph{test unseen}. For REVERIE, Table~\ref{tab:reverie_main} reports TL, OSR, SR, SPL, RGS, RGSPL, and runtime on \emph{val unseen}, as the official test server is unavailable. We treat SR, SPL, and RGSPL as primary task metrics; TL characterizes path efficiency, NE measures endpoint accuracy, and runtime reflects computational cost.

All variants use the same baseline checkpoints and differ only in their inference-time control settings; none requires retraining. Unless stated otherwise, DART-VLN denotes \emph{decay+anti-loop} under deterministic argmax inference. We report \emph{update-only}, \emph{decay-only}, and \emph{full-mode} as auxiliary ablations and stress tests. The default configuration uses \(\lambda=0.12\), \(\alpha=0.15\), \(w_{\min}=0.35\), \(w_{\max}=1.0\), \(\beta_{\mathrm{back}}=0.22\), \(\beta_{\mathrm{rev}}=0.06\), and \(k=2\). We use these hyperparameters in all DART-VLN comparisons without retraining or checkpoint selection. Write-side variants use \(g_0=0.15\), \(\lambda_n=0.35\), \(\lambda_a=0.20\), \(\lambda_c=0.15\), \(g_{\min}=0.05\), \(g_{\max}=0.85\), \(T_{\max}=20\), and \(r_{\mathrm{match}}=0.75\). Runtime is reported only for GridMM and its variants, evaluated with the same implementation on an NVIDIA GeForce RTX 5070 Ti; cross-paper runtime is not directly comparable. The reduction mainly reflects lower memory-access and downstream inference pressure, as decay concentrates readout on a distilled set of informative slots.

\subsection{Main Results on R2R and REVERIE}

Tables~\ref{tab:r2r_main} and~\ref{tab:reverie_main} show a consistent pattern: \emph{decay-only} preserves or improves task performance while reducing runtime, whereas \emph{decay+anti-loop} achieves the best balance among the GridMM variants. We discuss the main trends below and return to the stress-test variants in Sec.~\ref{sec:ablation_variants}.

\textbf{R2R.} Table~\ref{tab:r2r_main} reports results on both \emph{val unseen} and \emph{test unseen}. On this benchmark, \emph{decay-only} already shows the value of conservative read-side control: compared with GridMM, it improves NE from 2.83 to 2.59 on \emph{val unseen} and from 3.35 to 3.19 on \emph{test unseen}, while also reducing runtime from 937.99\,s to 742.61\,s and from 2312.52\,s to 1621.34\,s, respectively. Its SR and SPL are maintained or slightly improved, suggesting that read-side decay reduces the influence of stale memory while improving inference efficiency.

Among the GridMM variants, \emph{decay+anti-loop} achieves the best overall balance on R2R. It produces the shortest trajectories and lowest runtime on both splits while improving SR/SPL from 73/62 to 74/63 on \emph{test unseen}. Its test NE is slightly worse than GridMM (3.38 versus 3.35), so the result reflects a better balance between task performance and efficiency rather than a uniform improvement in endpoint accuracy.

\begin{table}[t]
\par\vspace{2pt}
\centering
{\footnotesize
\renewcommand{\arraystretch}{1.00}
\setlength{\tabcolsep}{1.8pt}
\renewcommand{\arraystretch}{1.15}
\caption{Results on the REVERIE val-unseen split.}
\vspace{-8pt}
\label{tab:reverie_main}
\begin{tabular}{c|cccc|cc|c}
\hline
\multirow{3}{*}{\textbf{Methods}} & \multicolumn{6}{c|}{\textbf{Val Unseen}} & \multirow{3}{*}{Runtime(s)$\downarrow$} \\
\cline{2-7}
 & \multicolumn{4}{c|}{\textbf{Navigation}} & \multicolumn{2}{c|}{\textbf{Grounding}} & \\
\cline{2-7}
 & TL$\downarrow$ & OSR$\uparrow$ & SR$\uparrow$ & SPL$\uparrow$ & RGS$\uparrow$ & RGSPL$\uparrow$ & \\
\hline
VLNBERT~\cite{hong2021vlnbert} & 16.75 & 35.08 & 30.63 & 24.90 & 18.72 & 15.23 & -- \\
AirBERT~\cite{guhur2021airbert} & 18.71 & 34.51 & 27.88 & 21.83 & 18.19 & 14.16 & -- \\
HOP~\cite{qiao2023hopplus} & 16.46 & 36.24 & 31.78 & 26.15 & 18.82 & 15.73 & -- \\
HAMT~\cite{chen2021hamt} & \textbf{14.08} & 36.84 & 32.91 & 30.23 & 18.90 & 17.26 & -- \\
TD-STP~\cite{zhao2022tdstp} & -- & 39.48 & 34.89 & 27.34 & 21.13 & 16.52 & -- \\
DUET~\cite{chen2022duet} & 22.12 & 51.08 & 46.96 & 33.77 & 32.12 & 23.01 & -- \\
BEVBert~\cite{an2022bevbert} & -- & 56.38 & 51.79 & 36.35 & 34.70 & 24.46 & -- \\
GridMM~\cite{wang2023gridmm}& 23.20 & 57.48 & 51.37 & 36.47 & 34.57 & 24.56 & 4329.67 \\
\hline
update-only & 23.77 & 57.12 & 50.84 & 35.85 & 33.98 & 23.80 & 4276.31 \\
decay-only & 23.15 & \textbf{58.12} & 51.98 & 36.60 & 34.68 & 24.72 & 2998.49 \\
full-mode & 23.18 & 57.94 & 51.80 & 36.45 & 34.65 & 24.68 & 4603.40 \\
\textbf{decay+anti-loop} & \textbf{21.57} & 57.99 & \textbf{52.34} & \textbf{37.53} & \textbf{35.37} & \textbf{25.44} & \textbf{1497.98} \\
\hline
\end{tabular}
\par\vspace{5pt}
\parbox{\columnwidth}{\scriptsize\emph{Note:} Runtime is reported only for GridMM and its variants under the same setting; the official test unseen server is unavailable.}
}
\par\vspace{-6pt}
\end{table}

\textbf{REVERIE.} Table~\ref{tab:reverie_main} shows the same qualitative trend on \emph{val unseen}. \emph{decay-only} slightly improves navigation and grounding metrics while reducing runtime, indicating that the benefits of read-side memory control extend to grounding-related metrics. Among the GridMM variants, \emph{decay+anti-loop} achieves the strongest overall performance, improving SR/SPL/RGSPL from 51.37/36.47/24.56 to 52.34/37.53/25.44 and reducing runtime from 4329.67\,s to 1497.98\,s. Together, the two benchmarks show that memory decay provides consistent improvements and that anti-loop regularization further reduces trajectory length and runtime.

\begin{figure*}[t]
\centering
\includegraphics[width=1.0\textwidth,height=5cm]{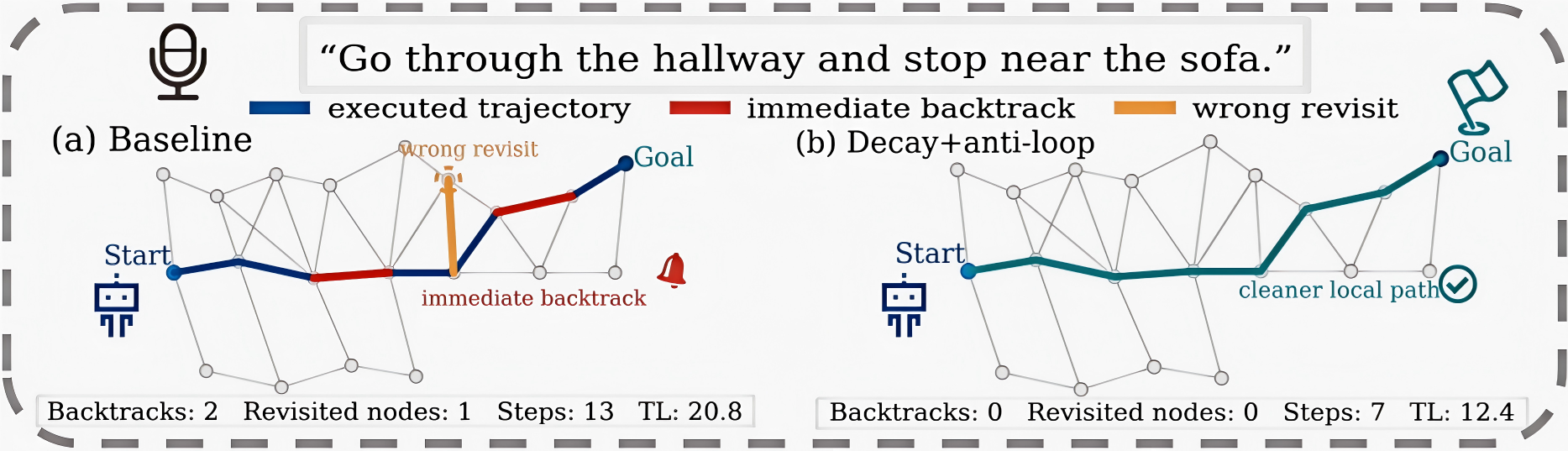}
\vspace{-10pt}
\caption{Example in which \emph{decay+anti-loop} avoids immediate backtracking and follows a shorter path.}
\label{fig:behavior_case}
\vspace{-10pt}
\end{figure*}

\subsection{Ablations and Stress-Test Variants}\label{sec:ablation_variants}

The comparison among \emph{update-only}, \emph{decay-only}, \emph{full-mode}, and \emph{decay+anti-loop} clarifies the role of each component. Among the single-module variants, \emph{decay-only} consistently preserves or improves task metrics while reducing runtime. By contrast, \emph{update-only} and \emph{full-mode} rewrite memory before readout and show less consistent gains, suggesting that aggressive write-side intervention can disturb representations already used by the frozen policy.

\emph{decay+anti-loop} builds on this read-side improvement with a narrow action-side controller that reduces local reversals, trajectory length, and runtime. The ablation supports the main design principle: with a frozen backbone, well-scoped test-time control is more reliable than heavier intervention.
\subsection{Navigation Behavior and Efficiency Analysis}

Anti-loop regularization is not meant to guarantee monotonic gains on all endpoint metrics; its direct value is local trajectory control. Figure~\ref{fig:behavior_case} illustrates this effect, while Table~\ref{tab:behavior_diag} quantifies it through backtracking rate, average action count, and trajectory length.

\begin{table}[t]
\small
\vspace{2pt}
\centering
\caption{Behavioral analysis on the R2R and REVERIE val-unseen splits.}
\vspace{-8pt}
\label{tab:behavior_diag}

\setlength{\tabcolsep}{1.6pt}
\renewcommand{\arraystretch}{1.08}
\begin{tabular}{c|ccc}
\hline
\multirow{2}{*}{\textbf{Methods}} & \multicolumn{3}{c}{{\small R2R Val Unseen}} \\
\cline{2-4}
& {\small Backtrack Rate(\%)$\downarrow$} & {\small Avg. Steps$\downarrow$} & {\small Avg. TL$\downarrow$}\tabularnewline
\hline
GridMM (baseline) & 2.30 & 6.02 & 13.27\tabularnewline
decay-only & 3.51 & 6.00 & 13.29\tabularnewline
\textbf{decay+anti-loop} & \textbf{2.01$\downarrow$} & \textbf{5.90$\downarrow$} & \textbf{12.41$\downarrow$}\tabularnewline
\hline
\end{tabular}

\vspace{4pt}

\begin{tabular}{c|ccc}

\hline

\multirow{2}{*}{\textbf{Methods}} & \multicolumn{3}{c}{{\small REVERIE Val Unseen}} \\
\cline{2-4}
& {\small Backtrack Rate(\%)$\downarrow$} & {\small Avg. Steps$\downarrow$} & {\small Avg. TL$\downarrow$}\tabularnewline
\hline
GridMM (baseline) & 8.43 & 8.52 & 23.20\tabularnewline
decay-only & 8.45 & 8.52 & 23.15\tabularnewline
\textbf{decay+anti-loop} & \textbf{5.99$\downarrow$} & \textbf{8.46$\downarrow$} & \textbf{21.57$\downarrow$}\tabularnewline
\hline
\end{tabular}

\par\vspace{4pt}
\parbox{\columnwidth}{\scriptsize\emph{Note:} Backtrack rate is the percentage of executed actions that return to the viewpoint from two steps earlier; Avg. Steps counts actions before STOP. Because the default configuration uses only a weak revisit penalty, the observed changes are dominated by immediate-backtracking suppression.}

\par\vspace{-5pt}
\end{table}

On R2R, \emph{decay-only} leaves local reversal behavior largely unchanged: its average action count changes only marginally, and its immediate-backtracking rate remains higher than that of the baseline. This is consistent with its role as a read-side denoising rule rather than a motion regularizer. By contrast, adding anti-loop regularization reduces the immediate-backtracking rate from 3.51\% to 2.01\% while also decreasing the average number of actions and trajectory length.

REVERIE exhibits the same pattern. The baseline and \emph{decay-only} remain nearly identical in local reversal behavior, whereas \emph{decay+anti-loop} reduces the backtracking rate from 8.45\% to 5.99\% and follows shorter trajectories. This consistency across benchmarks supports the intended role of anti-loop regularization as a graph-local controller rather than a general planning mechanism.

\section{Conclusion}

We have presented DART-VLN, a training-free plug-in control layer for discrete VLN pipelines with explicit memory. When instantiated on a GridMM-based navigator, DART-VLN improved navigation performance and efficiency on R2R and REVERIE without retraining or architectural changes. Read-side decay preserved or improved task performance while reducing runtime, whereas anti-loop regularization further reduced local backtracking and trajectory length. Together, these findings highlight the practical value of conservative test-time control for improving inference behavior and efficiency in frozen memory-based navigators.

\par\smallskip
Beyond the aggregate results, our behavioral analysis clarified how the two modules contributed at different stages of inference: decay stabilized memory readout, whereas anti-loop regularization improved local trajectory control. Their combination offers a compact and interpretable intervention for frozen navigators.

\par\smallskip
Our experiments focused on a GridMM-based instantiation, which provided a controlled setting for isolating these effects. Future work will examine transfer to other explicit-memory backbones and hyperparameter settings, as well as stochastic decoding and continuous navigation, to assess the broader applicability of conservative test-time control.






\let\oldthebibliography\thebibliography
\renewcommand{\thebibliography}[1]{%
  \oldthebibliography{#1}%
  \setlength{\itemsep}{0pt}%
  \setlength{\parskip}{2pt}%
}

\bibliographystyle{IEEEtran}
\bibliography{references}

\end{document}